\documentclass[journal]{IEEEtran}
\IEEEoverridecommandlockouts

\usepackage{cite}
\usepackage{amsmath,amssymb,amsfonts}
\usepackage{graphicx}
\usepackage{textcomp}
\usepackage{xcolor}
\usepackage{booktabs}
\usepackage{adjustbox}
\usepackage{float}
\usepackage{stfloats}
\usepackage{placeins}
\usepackage{subcaption}
\usepackage{caption}
\usepackage{threeparttable}
\usepackage{multirow}
\usepackage{colortbl}
\usepackage{hyperref}
\usepackage{pifont}
\usepackage[most]{tcolorbox}
\tcbset{promptbox/.style={
  breakable, enhanced jigsaw,
  colback=gray!5, colframe=black!60,
  fonttitle=\bfseries\footnotesize,
  sharp corners, boxrule=0.5pt,
  left=3pt, right=3pt, top=3pt, bottom=3pt,
  boxsep=1pt, before skip=5pt, after skip=5pt
}}
\definecolor{lightgreen}{RGB}{226,239,218}
\definecolor{lightred}{RGB}{255,224,224}
\definecolor{lightyellow}{RGB}{255,242,204}
\definecolor{lightblue}{RGB}{214,228,240}
\newcommand{\cmark}{\ding{51}}
\newcommand{\xmark}{\textcolor{black!35}{\ding{55}}}
\newcommand{\pmark}{\textcolor{black!55}{(\ding{51})}}   

\def\BibTeX{{\rm B\kern-.05em{\sc i\kern-.025em b}\kern-.08em
    T\kern-.1667em\lower.7ex\hbox{E}\kern-.125emX}}

\begin{document}

\title{Dissecting Neuro-Symbolic Quality Assurance for Synthetic Oncology Data Generation}
\author{
Laxmigayathri Challa, 
Yuhan Zhou, 
Ana Cleveland,
Haihua Chen,~\IEEEmembership{Member,~IEEE}
\thanks{L. Challa, Y. Zhou, and A. Cleveland are with the Department of Information Science, University of North Texas, Denton, TX 76203 USA.}
\thanks{H. Chen is with The Anuradha and Vikas Sinha Department of Data Science, University of North Texas, Denton, TX 76203 USA.}
\thanks{Corresponding author: Haihua Chen (e-mail: Haihua.Chen@unt.edu).}
}

\markboth{Manuscript Under Review}%
{Challa \MakeLowercase{\textit{et al.}}: Quality Assurance for Synthetic Oncology Data Generation}

\maketitle

\begin{abstract}
Synthetic clinical data generation with large language models addresses the
scarcity that limits cancer staging research, but oncology hallucinations are
categorically harmful: one clinically impossible staging assignment
contaminates every downstream model trained on it. Neuro-symbolic pipelines
validate during generation, yet the contribution of individual
quality-assurance components remains unclear. We report three controlled
studies isolating gate necessity, constraint attribution, and retrieval
conditionality, holding generation protocol, diversity thresholds, and
fine-tuning hyperparameters constant across adapter conditions. The symbolic
gate enforces schema completeness, ontology coverage against the Systematized
Nomenclature of Medicine, and staging-logic consistency under American Joint
Committee on Cancer eighth-edition rules. Ungated, 29.9\% of records contain
schema failures and 20.1\% contain clinically invalid staging. Schema
validation is the load-bearing filter: within the fully gated corpus it
rejects 148 of 512 records, ontology grounding a further 24, and staging-logic
validation none---the only generator producing logic violations is already
excluded on schema, making clinical-logic validation a generator-conditional
safeguard rather than the dominant filter. Retrieval augmentation is strongly
model-dependent: it improves gate compliance for one generator by 12.5
percentage points, has no measurable effect for a second, and collapses output
in a third. Across gated configurations ontology density is largely unchanged,
indicating that symbolic validation improves clinical validity rather than
vocabulary richness. Symbolic gating therefore buys corpus validity but no
commensurate gain on real lung-cancer notes in this study; retrieval should be
evaluated per model, and ontology density should not be reported as a proxy
for corpus quality.
\end{abstract} 


\section{Introduction}
\label{sec:intro}

Lung cancer remains the leading cause of cancer mortality worldwide, with an
estimated 2.48 million new cases and 1.8 million deaths in
2022~\cite{bray2024global}. Almost everything about a patient's prospects
turns on how far the disease has spread when it is found: in the United
States, five-year relative survival is 63.7\% while the tumor is still
localized and 8.9\% once it has metastasized, yet only 22\% of cases are
caught at the localized stage~\cite{seer2024lungb}. Stage decides whether a
patient is offered curative resection or systemic therapy, which trials they
are eligible for, and how their outcome is judged against expected care.

Facts of this kind are recorded for every patient, but they are recorded as
narrative (pathology reports, operative notes, oncology summaries) rather
than as coded fields a query can reach~\cite{hands2025survey}, and recovering
them is complex, time-consuming, and heavily
manual~\cite{tayefi2021challenges}.
Yet nearly every use we would make of
them operates at a scale prose cannot serve: assembling trial cohorts,
auditing care against guidelines, curating registries, and training the
clinical models meant to support all three. Working directly from real records runs into protections that exist for good
reason: patient data is governed under HIPAA and GDPR through slow,
jurisdiction-specific sharing mechanisms~\cite{price2019privacy}, and a
solution that works by loosening them is not a solution. Where records are
shareable, the clinically meaningful variables are rarely structured, and
recovering them retrospectively demands expertise no annotation budget
supports. Where they are reliably structured, the structure is thin. Registries capture
a fixed set of abstracted variables rather than the reasoning in the note;
trial data is richer but covers a narrow slice, with only 7.1\% of adults
with cancer enrolling in treatment trials~\cite{unger2024national}. The scarcity is structural,
and it will persist.

Synthetic generation offers a way through---clinical text carrying
the properties of real records while describing no real patient---and large
language models make it tractable, producing rich clinical narratives at
scale~\cite{chen2021synthetic}, showing promise across precision-oncology
tasks~\cite{liang2025precision}, and grounding in domain evidence through
retrieval-augmented generation (RAG)~\cite{lewis2020rag}. The difficulty is
knowing whether what comes out is correct, and for most clinical content
nothing decides that but expert judgment. Cancer staging is a deliberate
exception: under AJCC 8th Edition rules~\cite{ajcc2017}, T, N, and M values
map to a stage group through a fixed lookup, so a generated record either
satisfies the rules or contradicts them---machine-decidably, and without
reference to any patient. This work therefore begins with TNM staging:
clinically consequential in its own right, and structured enough that a
generator can be taught it and a validator can check it.
\begin{figure}[!t]
  \centering
  \includegraphics[width=0.5\textwidth]{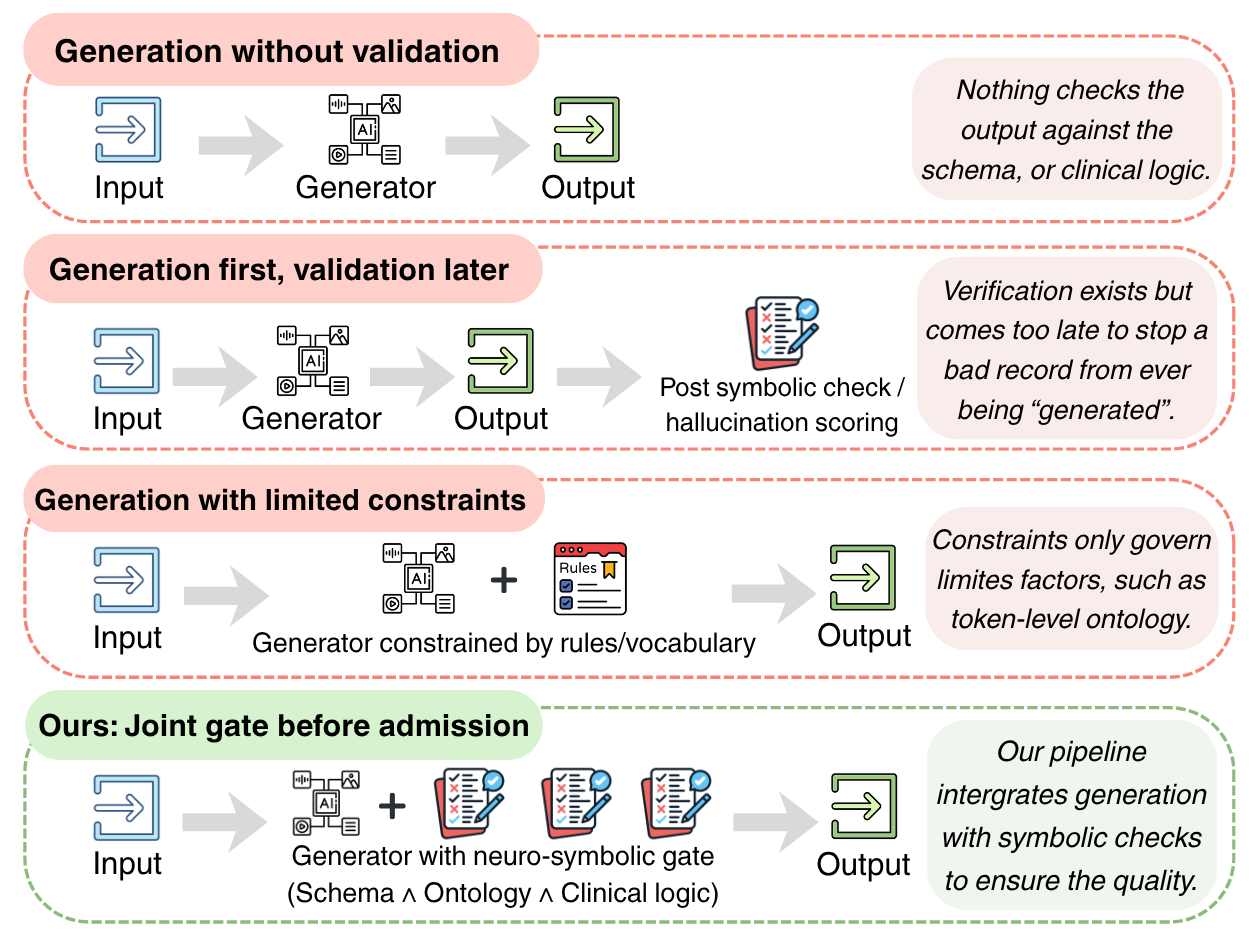}

  \caption{Demonstration of different synthetic data generation methods.}
  \label{fig:methodologyy}
\end{figure}

Feasible is not trustworthy. 
LLMs
hallucinate~\cite{huang2023hallucination}, and in oncology hallucination is
categorical rather than gradient: a note assigning T2\,N0\,M0 to mediastinal
lymph node involvement is not slightly off but clinically impossible, and any
model trained on it inherits an error that propagates without warning.
Model-collapse work~\cite{shumailov2024collapse} sharpens the concern, since
generative systems trained on their own outputs degrade across generations.
Corpus quality therefore cannot be repaired downstream; it has to be enforced
at the point of generation. The natural response is a symbolic gate:
deterministic validators that screen every record before it enters the
corpus, enforcing JSON schema completeness, SNOMED\,CT ontology coverage, and
AJCC clinical-logic consistency as hard admission constraints. Prior
work~\cite{priorpipeline} has shown that such a pipeline---neuro-symbolic
generation coupled with a binary gate $G(x) = S \wedge O \wedge C$---improves
downstream T-stage extraction on real clinical notes, recovering minority
stages that zero-shot models cannot identify at all. That establishes the
pipeline as a whole. It does not establish \emph{which components produce
the effect}, nor whether the quality signals used to justify them measure
what they are assumed to measure.
 
The purpose of this study is to quantify the marginal
contribution of each quality-assurance component---the gate as a whole, each
of its three constraints ($S$, $O$, $C$), and retrieval augmentation---to two
distinct outcomes: the clinical validity of the generated corpus, and its
downstream utility on real oncology notes under a train-on-synthetic,
test-on-real protocol. Each of these components carries a real cost in API calls,
computational overhead, reduced yield, and architectural complexity, and a
practitioner needs to know which of those costs are justified. A system study
answers whether an architecture works; only a controlled decomposition
answers which of its parts are load-bearing, and at what cost.

The research questions below are operationalized through controlled experiments in which a single pipeline component is manipulated while all others are held constant. Generation protocols, label diversity, entropy thresholds, QLoRA hyperparameters, and evaluation procedures are fixed across conditions, ensuring that observed differences are attributable to the manipulated variable alone. The research questions are:
\begin{enumerate}

\item \textbf{RQ1 (Gate Necessity):}
What enters a synthetic clinical corpus when symbolic quality assurance is removed?


\item \textbf{RQ2 (Constraint Attribution):}
Which symbolic constraint contributes most to corpus quality?


\item \textbf{RQ3 (Retrieval Conditionality):}
When does retrieval augmentation improve generation quality?

Section~\ref{sec:framework} maps each
question to the conditions that answer it.


\end{enumerate}
The contributions of this study are as follows:

\begin{itemize}
  \item \textbf{A component attribution methodology:} Five matched adapter
  conditions and a within-corpus decomposition procedure that isolate the
  gate, its three constraints, and retrieval as independently testable
  variables under fixed generation and fine-tuning protocols.

  \item \textbf{Evidence on what gating does and does not buy:} The
  failures an ungated corpus admits are structurally indistinguishable from
  valid records, establishing a class of contamination that post-hoc
  inspection cannot recover---and, separately, that removing it does not by itself improve transfer to real pathology reports.

  \item \textbf{A ranking of the symbolic constraints by marginal
  contribution:} We separate filtering work from standardization work, and
  identify clinical-logic validation as a generator-conditional safeguard
  rather than a high-volume filter.

  \item \textbf{A per-generator decision procedure for retrieval:} Gate
  compliance serves as the diagnostic for whether retrieval will help, do
  nothing, or cause output collapse in a given generator.
\end{itemize}

\section{Related Work}
\label{sec:related}
Across synthetic clinical data generation, hallucination research, retrieval-
augmented generation, and neuro-symbolic clinical NLP, validation is
consistently applied \emph{after} generation: as a benchmark score, a
post-hoc filter, or a normalization pass over records already produced. None
of these literatures treats the validators themselves as a variable to be
measured, so a pipeline that couples generation with symbolic checks is
evaluated only end-to-end---leaving the marginal contribution of any single
quality-assurance component unknown. Table~\ref{tab:related} situates the
present work against the closest prior systems on this dimension.
\subsection{Synthetic Clinical Data: From Tabular Simulation to LLM Generation}
\label{sec:rel_synth}

The earliest synthetic clinical data systems operated on structured EHR
extracts. GAN-based approaches---MedGAN~\cite{choi2017medgan},
medBGAN~\cite{baowaly2019synthesizing}, and
HealthGAN~\cite{yale2020generation}---learn to approximate the marginal
and joint statistics of tabular records and have been validated through
TSTR protocols on structured data~\cite{gonzales2023synthetic}.
Synthea~\cite{walonoski2018synthea} takes a mechanistic route, producing
structurally valid FHIR records by construction from probabilistic disease
models. Both paradigms share a structural ceiling: their output is a row of
feature values, and the unstructured clinical text in which a physician
encodes staging rationale---tumor invasion depth, lymph node involvement,
metastatic spread---is outside their generative scope entirely. Ontologies, where they appear at all, serve as vocabulary mappings applied to
already-generated content rather than as constraints on what enters the
corpus.

LLM-based generation removed the free-text barrier~\cite{tang2023synthetic,
kasthurirathne2023synthetic}. Models conditioned on structured metadata
produce linguistically plausible clinical notes at scale, and
MedSyn~\cite{kumichev2024medsyn} demonstrated that knowledge-graph-conditioned
generation can improve downstream {ICD}-10 coding accuracy, with the largest
gains on long-tail codes. Yet the dominant
design pattern carries over from the GAN era unchanged: records are
produced first, and quality checks follow. Ontology mapping tools---
MetaMap~\cite{aronson2010metamap}, cTAKES~\cite{savova2010mayo}, and the
BioPortal Annotator API~\cite{whetzel2011bioportal}---are applied as
normalization steps on records that already exist---so post-hoc validation cannot catch what it cannot see.

This work enforces ontology coverage and clinical-logic consistency as
admission constraints \emph{during} generation rather than as normalization
applied afterward (Fig.\ref{fig:methodologyy}).

\subsection{Hallucination in Clinical LLMs: Evaluation Versus Prevention}
\label{sec:rel_halluc}
Hallucination in clinical LLMs has so far been treated chiefly as an
evaluation problem rather than a prevention one. Med-HALT~\cite{pal2023medhalt}
introduced a benchmark taxonomy for medical hallucination types; large-scale
clinical LLM evaluations~\cite{singhal2023large} showed that high scores on
general benchmarks do not transfer reliably to domain-specific clinical
tasks requiring precise factual grounding; MedHallu~\cite{pandit2024medhallu}
extended this to a comprehensive detection benchmark across clinical
domains. Together these establish that hallucination is pervasive,
measurable, and domain-specific---but measuring it after the fact is not
the same as preventing it at the point of generation.

The clinical hallucination analysis of~\cite{asgari2025framework} applies
clinician-in-the-loop evaluation as post-hoc analysis, correctly identifying
that hallucination rates differ by task type and model family, but again
after generation is complete. The closest prior approach to generation-time
prevention is ontology-guided constrained decoding~\cite{mehenni2024ontology},
which reduces hallucination during clinical summarization by restricting the
model's output vocabulary to ontology-consistent tokens. Constrained
decoding, however, operates on individual token predictions; it does not
enforce higher-level semantic constraints such as AJCC staging logic, which
requires cross-field consistency across an entire structured record rather
than local token validity.

That gap reflects a distinction this paper treats as load-bearing:
hallucination as a \emph{gradient} failure---a claim that is somewhat
inaccurate but recoverable---versus hallucination as a \emph{categorical}
failure, one that is clinically impossible and corrupts downstream learning
silently. In oncology the latter dominates. Embedding AJCC rules as a
binary gate is what converts this categorical failure from an invisible
contaminant into a measurable engineering property, and the gate
decomposition experiment in this paper quantifies how much of that
filtering work the logic constraint performs relative to schema and
ontology.
\subsection{Neuro-Symbolic Approaches in Clinical NLP}
\label{sec:rel_neurosym}

The neuro-symbolic paradigm~\cite{garcez2022neural} combines the generative
expressiveness of neural models with the precision of symbolic reasoning,
and in clinical NLP it has been applied chiefly to classification and
coding rather than generation. Hybrid-Code~v2~\cite{yu2025hybridcode} uses
neuro-symbolic verification for ICD-10 coding, achieving zero Type-I
hallucination on MIMIC-III by layering symbolic rule checks over neural
candidate generation. Ontology-guided decoding~\cite{mehenni2024ontology},
introduced above as the closest prior approach to generation-time
constraint enforcement, is the same pattern applied to summarization. In
both cases the symbolic component operates as a post-generation filter or
decoding constraint, not as a corpus admission gate, and neither addresses
synthetic data \emph{generation} at all.

The architectural distinction between a filter and a gate is not merely
terminological. A post-generation filter incurs the full cost of generating
every record before deciding which to discard. A corpus admission gate
makes admission a binary property of each record independently: records
that fail do not enter the corpus, and the gate's pass rates are directly
interpretable as measurements of pipeline component contribution. This
interpretability is what the component attribution study in this paper
exploits---by varying which constraints are active while holding all other
variables fixed, the marginal downstream contribution of each symbolic
component becomes directly measurable rather than inferred.

\subsection{Retrieval-Augmented Generation in Biomedicine}
\label{sec:rel_rag}

Retrieval-augmented generation (RAG)~\cite{lewis2020rag} grounds LLM outputs in
verifiable retrieved evidence, reducing hallucination by anchoring
generation to specific source documents. MedCPT~\cite{jin2023medcpt}
extends this to biomedicine through contrastive pre-training on PubMed
citation pairs, enabling zero-shot biomedical retrieval without
task-specific fine-tuning. Clinical LLM work has concentrated on question
answering~\cite{singhal2023large} and summarization, and where retrieval has
been applied it has consistently improved factual accuracy---and it is this consistency that
has hardened into an implicit assumption: that retrieval is universally
beneficial, and more grounding context simply produces more accurate
output.

To our knowledge, that assumption has not been tested in the structured generation setting
used here, where the interaction between retrieved context and a model's
format-compliance capability is non-trivial. A model with fragile
instruction-following may fail to integrate retrieved content without
format disintegration, reducing effective corpus yield even as the
retrieved context enriches the individual records that do pass validation.
The retrieval ablation in this paper is the first controlled test of the
universality assumption in a structured clinical data generation context,
and the results challenge it directly: retrieval improves one generator,
has no measurable effect on a second, and causes systematic output collapse
in a third.

\subsection{TSTR Evaluation and TNM Staging Automation}
\label{sec:rel_tstr}

The Train-on-Synthetic, Test-on-Real (TSTR)
protocol~\cite{esteban2022longitudinal}
measures synthetic data utility through downstream task performance rather
than distribution-similarity proxies.
Prior clinical TSTR evaluations have focused on structured
tabular EHR data~\cite{yale2020generation,gonzales2023synthetic}; this
paper applies TSTR to synthetic oncology note generation with explicit
staging annotations, evaluated on TCGA-derived lung-cancer and cross-tumor
clinical note cohorts.

Cancer staging extraction spans rule-based methods, classical machine
learning, and deep learning in comparable numbers, with transformer-based
approaches a smaller but growing share~\cite{hands2025survey}. Reported performance on curated benchmarks is strong---macro-F1 above 0.90
for both a transformer-plus-rules staging pipeline~\cite{hu2021staging} and a
fine-tuned generative model applied to the same
task~\cite{kim2024pathologic}---but performance on real-world oncology notes
remains substantially lower, a gap that reflects the data rather than the
model:
curated benchmarks provide clean, expert-annotated records, while clinical
dictation is noisy, abbreviated, and inconsistently formatted. Synthetic
data generation targeted at the structured oncology domain is therefore a
direct intervention on the most limiting constraint in TNM staging
automation. Whether the symbolic quality-assurance components of such a
pipeline are individually necessary to realize that intervention, and which
contribute most, is the question the remainder of this paper answers.

\begin{table*}[!t]
\centering
\caption{Capability comparison with representative prior work on synthetic
clinical data generation, hallucination prevention, and neuro-symbolic
medical NLP. 
Per-system detail is given
in Section~\ref{sec:related}.}
\textbf{Legend:} \cmark~present; \pmark~present but not as a generation-time,
record-level admission constraint; \xmark~absent. 
\label{tab:related}
\renewcommand{\arraystretch}{1.3}
\setlength{\tabcolsep}{5pt}
\footnotesize
\begin{threeparttable}
\begin{tabular}{@{}
  >{\raggedright\arraybackslash}p{4.0cm}
  >{\raggedright\arraybackslash}p{3.9cm}
  *{6}{>{\centering\arraybackslash}p{1.15cm}}@{}}
\toprule
 & & & \multicolumn{3}{c}{\textbf{Neuro-symbolic gate}} & & \\
\cmidrule(lr){4-6}
\textbf{Work} & \textbf{System type} &
\textbf{Free-text} & \textbf{Schema} & \textbf{Ontology} &
\textbf{Clinical logic} & \textbf{Gen-time gate} & \textbf{Attribution} \\
\midrule
MedGAN (2017)~\cite{choi2017medgan}
  & Tabular EHR (GAN) & \xmark & \xmark & \xmark & \xmark & \xmark & \xmark \\
Synthea (2018)~\cite{walonoski2018synthea}
  & FHIR record sim.\ (rule-based) & \xmark & \pmark\tnote{a} & \xmark & \xmark & \xmark & \xmark \\
MedSyn (2024)~\cite{kumichev2024medsyn}
  & LLM text (KG-conditioned) & \cmark & \xmark & \pmark\tnote{b} & \xmark & \xmark & \xmark \\
Kasthurirathne et al.\ (2023)~\cite{kasthurirathne2023synthetic}
  & LLM text & \cmark & \xmark & \xmark & \xmark & \xmark & \xmark \\
Mehenni \& Zouaq (2024)~\cite{mehenni2024ontology}
  & Constrained decoding & \cmark & \xmark & \cmark & \xmark & \pmark\tnote{c} & \xmark \\
Asgari et al.\ (2025)~\cite{asgari2025framework}
  & Hallucination eval & \cmark & \xmark & NA & \xmark & \xmark & \xmark \\
\midrule
\rowcolor{lightblue}
\textbf{Our work (2026)}
  & \textbf{Neuro-symbolic LLM + TSTR}
  & \cmark & \cmark & \cmark & \cmark & \cmark & \cmark \\
\bottomrule
\end{tabular}
\begin{tablenotes}[flush]
\footnotesize
\item[a] FHIR records are structurally valid \emph{by construction}, not validated as a generation-time admission gate.
\item[b] Knowledge-graph sampling conditions the generation prompt; it biases output toward valid content but rejects nothing.
\item[c] Constraints act at generation time but on token-level vocabulary, not as a record-level admission gate.

\end{tablenotes}
\end{threeparttable}
\end{table*}

\section{Clinical Foundation}
\label{sec:background}

\subsection{TNM Staging and AJCC 8th Edition}
\label{sec:bg_tnm}

The TNM classification system codifies cancer extent along three independent
dimensions. $T$ (T0--T4) describes the size and local invasiveness of the
primary tumor; $N$ (N0--N3) describes regional lymph node involvement; $M$
(M0/M1) describes the presence or absence of distant metastasis. Combined
under AJCC 8th Edition rules~\cite{ajcc2017}, these three values determine
a clinical stage group (I--IV) through a deterministic lookup: the mapping
is fixed, not probabilistic, and violations are logically impossible rather
than merely unlikely.

This determinism is consequential for synthetic data generation. A generated
record assigning M1 staging without a corresponding Stage~IV designation, or
Stage~I with N2 lymph node involvement, is not an imprecise approximation
of a valid record---it is a logical contradiction. Such contradictions pass
structural formatting checks, carry normal ontology density, and satisfy
JSON schema requirements. They are invisible to every quality metric except
an explicit clinical-logic validator. Identifying and quantifying how many
generated records contain such contradictions---and how much downstream
harm they cause when admitted to a training corpus---is one of the central
questions this paper answers.

For non-small cell lung cancer, the 32-cell grid
$\{$T1, T2, T3, T4$\} \times \{$N0, N1, N2, N3$\} \times \{$M0, M1$\}$
covers the main staging categories. Within this grid, T3 and T4 carry particular clinical weight: they mark the
point at which local tumor extent begins to determine resectability and the
choice between surgery and definitive chemoradiation, and they are the
categories a staging model most often gets wrong. A model that cannot
distinguish T3 from T4 is least reliable exactly where the treatment
decision is most consequential. T3 and T4 are also minority classes in real cohorts, which makes macro-F1
rather than aggregate accuracy the informative measure of staging performance
on the real-world oncology note cohorts evaluated here.

\subsection{Clinical Ontologies as Validity Anchors}
\label{sec:bg_ontology}
The Systematized Nomenclature of Medicine---Clinical Terms (SNOMED\,CT)
provides a standardized, hierarchically organized vocabulary of more than
360,000 clinical concepts~\cite{snomed2026}, covering anatomy, findings,
procedures, substances, and observable entities.
Its relevance to synthetic data quality is not as a retrieval resource but
as a \emph{validity anchor}: a generated clinical record whose concepts
cannot be mapped to any SNOMED\,CT term is producing fabricated
terminology---terminology that has no grounding in any recognized clinical
vocabulary. This is detectable without access to the patient's actual
record, making SNOMED\,CT coverage checkable on the generated text alone

We operationalize this through the BioPortal Annotator
API~\cite{whetzel2011bioportal}, which maps free text to SNOMED\,CT
concepts and returns matched terms with concept IDs. Length-normalized
SNOMED density---terms per 100 words---provides a cross-model vocabulary
richness metric used throughout the ablation analyses. A key question this
paper addresses is whether enforcing SNOMED\,CT coverage as a corpus
admission constraint produces records with measurably richer clinical
vocabulary, or whether the gate targets validity rather than richness. The
gate decomposition experiment answers this directly.

\section{Methodology}
\label{sec:framework}
This section describes the neuro-symbolic generation pipeline and the
experimental design through which its components are isolated. The pipeline
couples LLM generation with a symbolic gate $G(x) = S \wedge O \wedge C$
applied to every candidate record before admission to the training corpus;
in the retrieval condition, MedCPT-retrieved PubMed context is prepended to
the generation prompt before the gate is applied.
Figure~\ref{fig:methodology} presents the full pipeline (top) together with
the five matched conditions through which its components are dissected
(bottom).

\subsection{Problem Formulation}
\label{sec:formulation}

We formulate synthetic clinical data generation as a constrained generation
problem. Let $x$ denote a candidate structured patient record produced by a
generator model $\mathcal{M}$ from a prompt $p$:
$x \sim \mathcal{M}(\theta, p)$, where $\theta$ denotes model parameters.
For $x$ to be admitted to the training corpus, it must satisfy a constraint
set $\mathcal{C} = \{c_{\text{schema}}, c_{\text{ontology}},
c_{\text{logic}}\}$, evaluated through a binary symbolic gate:

\begin{equation}
  G(x) =
  \begin{cases}
    1 & \text{if } x \text{ satisfies all } c \in \mathcal{C}, \\
    0 & \text{otherwise.}
  \end{cases}
  \label{eq:gate}
\end{equation}

The gate decomposes as $G(x) = S \wedge O \wedge C$, where $S$ denotes
schema validity (JSON field completeness enforced by the \textit{rigid.v3}
schema), $O$ denotes ontology coverage (at least one SNOMED\,CT concept
confirmed via the BioPortal Annotator API~\cite{whetzel2011bioportal}), and
$C$ denotes AJCC clinical-logic consistency, implemented over the most
commonly violated rules (M1 staging implies Stage~IV; stage group is
non-empty when T, N, and M values are all present); Section~\ref{sec:res_abl2}
reports the effect of expanding this rule set. The gate
is applied per record during corpus construction: records for which
$G(x) = 0$ are discarded and do not enter the fine-tuning corpus under any
condition.

Pipeline quality is assessed along three dimensions: structural validity
(JSON schema compliance rate), label diversity (Shannon entropy per TNM
dimension, with floors $H_{T}, H_{N} \geq 1.109$ and $H_{M} \geq 0.554$),
and downstream utility (T-, N-, and M-stage extraction accuracy and macro-F1
under TSTR).

\subsection{Generation Setup}
\label{sec:generation}

\begin{figure*}[!t]
  \centering
  \includegraphics[width=0.8\textwidth]{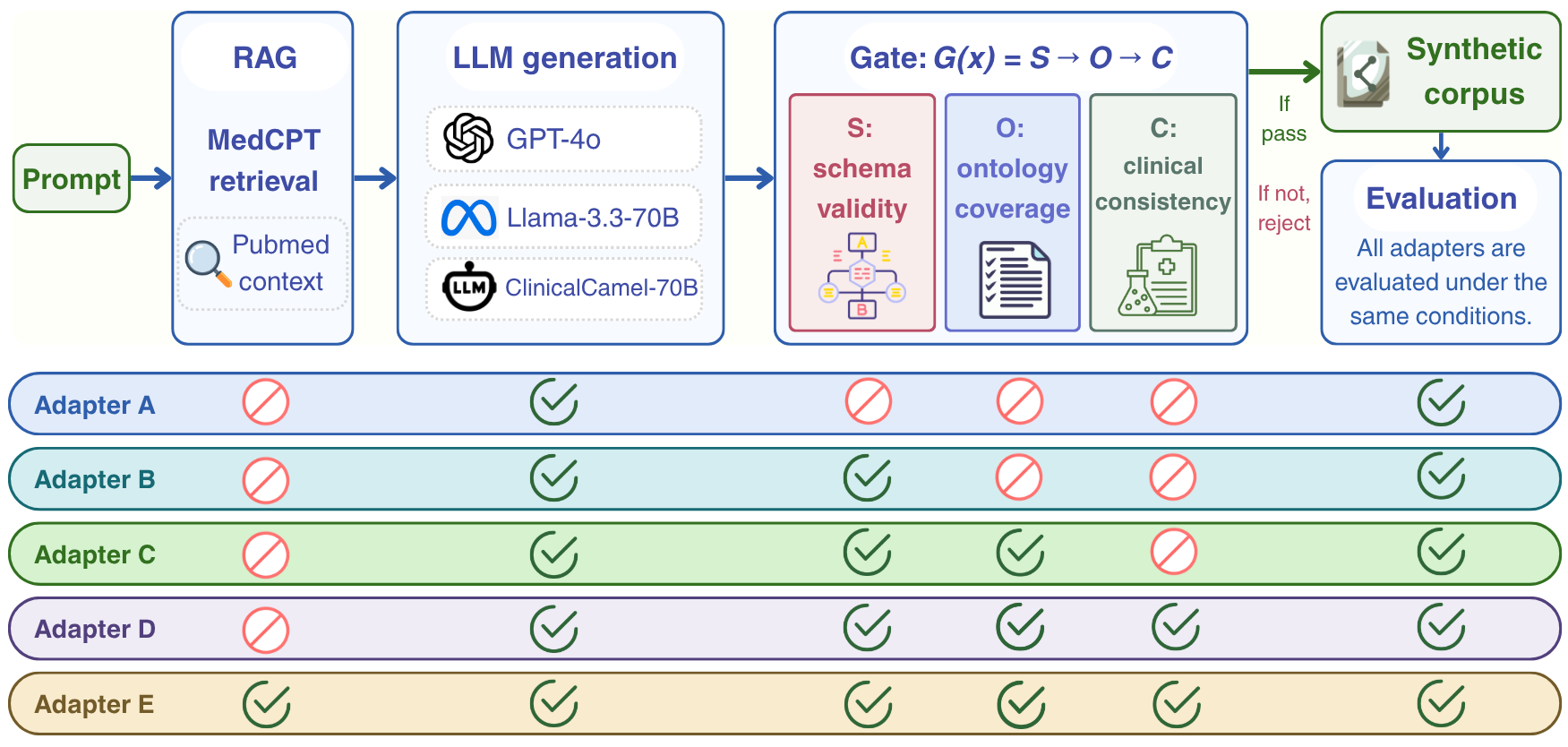}

  \caption{\textbf{Neuro-symbolic generation pipeline and its component
dissection.} \emph{Top (target pipeline).} The full framework couples LLM
generation with a neuro-symbolic gate $G(x) = S \wedge O \wedge C$---schema
validity ($S$), SNOMED\,CT ontology coverage ($O$), and AJCC clinical-logic
consistency ($C$)---and, in the retrieval condition, prepends
MedCPT-retrieved PubMed context to the generation prompt as grounding before
the gate is applied. Records that pass the gate enter the synthetic corpus
as structured \textit{rigid.v3} fields plus a free-text clinical narrative.
\emph{Bottom (dissection of components).} Five matched conditions
(Table~\ref{tab:adapters}) remove or add one component at a time: ungated
baseline (Adapter~A); schema only (Adapter~B); schema$+$ontology
(Adapter~C); full gate, no retrieval (Adapter~D); and full gate$+$RAG
(Adapter~E). The three within-gate conditions (B,~C,~D) isolate the marginal
contribution of each constraint (RQ~2); A versus D isolates gate necessity
(RQ~1); and D versus E isolates retrieval conditionality (RQ~3). The three
generators (GPT-4o, Llama-3.3-70B, ClinicalCamel-70B), the 32-cell TNM
seeding grid, the entropy floors, the QLoRA hyperparameters, and the TSTR
evaluation protocol are held fixed across every condition
(Sections~\ref{sec:generation}--\ref{sec:evalprotocol}), so that any
measured difference is attributable to the manipulated component alone.}
  \label{fig:methodology}
\end{figure*}

All ablation conditions share the generation setup described below.

\textbf{TNM grid:} As discussed in Section~\ref{sec:bg_tnm}, records are
seeded from a 32-cell TNM grid $\{$T1, T2, T3, T4$\} \times \{$N0, N1, N2,
N3$\} \times \{$M0, M1$\}$, assigned round-robin across generation runs,
with each cell randomizing patient age (45--80 years), sex, and histology
subtype.

\textbf{Label diversity:} Two Shannon-entropy checkpoints, applied
post-generation and pre-training, abort the pipeline if any TNM dimension
approaches single-class concentration. Floors and checkpoint placement are
given in Appendix~\ref{app:diversity}.

\textbf{Schema:} The \textit{rigid.v3} schema encodes eight non-optional
clinical domains per record: TNM staging (AJCC 8th Edition), histology
(ICD-O-3 and SNOMED\,CT codes), molecular drivers, demographics, imaging
findings, treatment modalities, health equity factors, and free-text
narrative. All fields are mandatory; empty strings trigger $G(x) = 0$.

\textbf{Generator models:} Three models are evaluated across all ablation
conditions. \textbf{GPT-4o}~\cite{openai2023gpt4} is accessed via the
OpenAI API with \texttt{response\_format=\{type: json\_object\}}, which
enforces JSON output at the API level.
\textbf{Llama-3.3-70B-Instruct}~\cite{grattafiori2024llama3} is loaded
locally in 4-bit NF4 quantization on an NVIDIA H200 GPU.
\textbf{ClinicalCamel-70B}~\cite{toma2023clinical} is loaded under the same
quantization. It does not ship with a chat template, so we attach a
Llama-style \texttt{<|start\_header\_id|>} fallback template at model load
time; this is the only model-specific modification applied in any condition.
All three models receive identical prompts per ablation condition, and
identical decoding parameters throughout (temperature 0.7, top-$p$ 0.9,
top-$k$ 50, sampling enabled, 1{,}024 maximum new tokens).

\textbf{Retrieval:} In Ablation~3 (RAG condition only), each generation call
is preceded by a MedCPT Query Encoder~\cite{jin2023medcpt} lookup over a
FAISS~\cite{douze2024faiss} flat inner-product index built from
${\approx}2{,}000$ PubMed lung-cancer abstracts encoded with the MedCPT
Article Encoder. The query is constructed from the record's T, N, and M seed
values together with its histology subtype; the top $k=2$ retrieved
abstracts are prepended to the generation prompt as grounding context. In
all other conditions---including the no-RAG arm of Ablation~3---no retrieval
context is added.
\footnote{Generation calls fall back to a keyword retriever when the MedCPT
index returns no qualifying match: 64 of 448 calls (32 Llama-3.3-70B and 32
ClinicalCamel-70B); all GPT-4o calls use MedCPT.}

The component attribution study rests on a single design principle: one
variable is manipulated per experiment while all others are held fixed.
Table~\ref{tab:adapters} summarizes the five adapter conditions and their
mapping to the three ablation experiments, corresponding to the dissection
panel of Figure~\ref{fig:methodology}. Each adapter is trained on a
condition-specific corpus generated under the corresponding study
configuration.
\begin{tcolorbox}[
  enhanced jigsaw, breakable,
  colback=black!3, colframe=black!60,
  boxrule=0.4pt, arc=2pt,
  left=5pt, right=5pt, top=4pt, bottom=4pt,
  title={\textbf{Research questions and how they are operationalized}},
  fonttitle=\footnotesize, coltitle=black, colbacktitle=black!12
]
\footnotesize
\setlength{\tabcolsep}{3pt}
\renewcommand{\arraystretch}{1.2}
\begin{tabular}{@{}p{0.55cm}p{3.5cm}p{1.7cm}p{1.5cm}@{}}
\textbf{RQ} & \textbf{Question} & \textbf{Contrast} & \textbf{Answered in} \\[2pt]
\hline\\[-6pt]
1 & Gate necessity: what enters a corpus when symbolic quality assurance is removed?
  & A vs.\ D
  & \S\ref{sec:abl1}, \S\ref{sec:res_abl1} \\
2 & Constraint attribution: which symbolic constraint does the filtering work?
  & B, C, D; per-record $S$, $O$, $C$ flags
  & \S\ref{sec:abl2}, \S\ref{sec:res_abl2} \\
3 & Retrieval conditionality: when does retrieval augmentation improve quality?
  & D vs.\ E
  & \S\ref{sec:abl3}, \S\ref{sec:res_abl3} \\
\end{tabular}

\vspace{3pt}
\noindent Generation protocol, TNM seeding grid, entropy floors, QLoRA
hyperparameters, and evaluation procedure are fixed across all conditions,
so each contrast varies one component alone.
\end{tcolorbox}
\begin{table}[!t]
\centering
\caption{Adapter conditions and ablation mapping. Adapter~D is the shared
full-gate, no-RAG baseline across all three ablations. Per-model run counts differ by construction: GPT-4o contributes 128
generations in every condition, 64 API calls having returned no output;
Llama-3.3-70B and ClinicalCamel-70B contribute 192 each (no-RAG) and 160 each
(RAG). Totals are 512 per no-RAG condition and 448 for the RAG condition.
All adapters use identical QLoRA configuration (Appendix~\ref{app:qlora}).}
\label{tab:adapters}
\begin{tabular}{@{}clcc@{}}
\toprule
\textbf{Adapter} & \textbf{Gate condition} & \textbf{RAG} & \textbf{Ablation} \\
\midrule
A & None (ungated)              & No  & 1 \\
B & $S$ only                    & No  & 2 \\
C & $S \wedge O$                & No  & 2 \\
D & $S \wedge O \wedge C$       & No  & 1, 2, 3 \\
E & $S \wedge O \wedge C$       & Yes & 3 \\
\bottomrule
\end{tabular}
\end{table}
\subsection{Ablation Design}
\label{sec:ablation_design}
\subsubsection{Ablation 1: Gate Necessity (RQ1)}
\label{sec:abl1}

Two matched corpora are generated from the same TNM grid using the same
models and hyperparameters. The \emph{ungated} corpus (Adapter~A) admits
every record for which JSON parses, regardless of schema completeness,
SNOMED\,CT coverage, or AJCC logic. The \emph{gated} corpus (Adapter~D)
admits only records passing the full gate $G(x) = S \wedge O \wedge C$. Both
corpora must pass all three entropy floors before training. Gate component
flags ($S$, $O$, $C$) are logged for every record in the ungated corpus
whether or not they are used for admission.

\subsubsection{Ablation 2: Constraint Attribution (RQ2)}
\label{sec:abl2}

Three corpora are generated under progressively stricter gates:

\begin{enumerate}
  \item \textbf{Schema only ($S$):} Admits records passing JSON completeness;
  SNOMED\,CT and AJCC logic are computed and logged but not used for
  admission. Trains Adapter~B.
  \item \textbf{Schema + Ontology ($S \wedge O$):} Adds SNOMED\,CT coverage
  as an admission requirement; AJCC logic is computed and logged but not
  enforced. Trains Adapter~C.
  \item \textbf{Full gate ($S \wedge O \wedge C$):} All three constraints
  enforced simultaneously. Trains Adapter~D.
\end{enumerate}

Although three corpora are generated to train Adapters~B--D, the marginal
contribution of each constraint is computed \emph{within a single corpus}
from the per-record gate flags ($S$, $O$, $C$ logged for every generated
record), so that the attribution reflects the constraints themselves rather
than sampling variation between independently generated corpora. SNOMED
density is computed across all three conditions on the admitted-record
basis.

\subsubsection{Ablation 3: Retrieval Conditionality (RQ3)}
\label{sec:abl3}

Two matched corpora, identical in every respect except retrieval context.
Both pass the full gate $G(x) = S \wedge O \wedge C$ and the same entropy
floors. The \emph{no-RAG} corpus (Adapter~D) uses schema plus TNM seed
description only. The \emph{RAG} corpus (Adapter~E) prepends
MedCPT-retrieved PubMed abstracts to each generation prompt via the FAISS
index described in Section~\ref{sec:generation}. Both corpora are generated
from the same TNM grid with the same three models.

\subsection{Evaluation Protocol}
\label{sec:evalprotocol}

All five adapters are evaluated under identical TSTR conditions at three
levels of distributional distance from the training distribution:

\begin{enumerate}
\item \textbf{Synthetic held-out} ($n=62$):
in-distribution evaluation.

\item \textbf{TCGA Lung} ($n=737$):
real lung-cancer notes.

\item \textbf{TCGA Cross-Tumor} ($n=3{,}161$):
out-of-distribution oncology evaluation.
\end{enumerate}

For the evaluation, accuracy is reported against the majority-class baseline,
with macro-F1 as the metric sensitive to minority-class recovery. Per-axis scoring
is restricted to notes carrying a valid gold label for that axis, and
per-axis macro-F1 is reported in Appendix~\ref{app:macrof1}.

The base model for all adapters is
\texttt{meta-llama/Meta-Llama-3-8B-Instruct}; full QLoRA hyperparameters
are given in Appendix~\ref{app:qlora}.
\section{Results}
\label{sec:results}
\begin{tcolorbox}[
  enhanced jigsaw, breakable,
  colback=black!3, colframe=black!60,
  boxrule=0.4pt, arc=2pt,
  left=5pt, right=5pt, top=4pt, bottom=4pt,
  title={\textbf{Findings by research question}},
  fonttitle=\footnotesize, coltitle=black, colbacktitle=black!12
]
\footnotesize
\setlength{\tabcolsep}{3pt}
\renewcommand{\arraystretch}{1.2}
\begin{tabular}{@{}p{0.4cm}p{4.7cm}p{1.9cm}@{}}
\textbf{RQ} & \textbf{Finding} & \textbf{Evidence} \\[2pt]
\hline\\[-6pt]
1 & Ungated generation admits 29.9\% schema failures and 20.1\% AJCC logic
    violations; the full gate rejects one record in three, and the cost falls
    almost entirely on a single generator.
  & Fig.~\ref{fig:abl1}, Tables~\ref{tab:corpus_quality},
    \ref{tab:abl1_model} \\
2 & Schema is the load-bearing filter (148 of 512 rejected); ontology removes
    24 more; clinical logic removes none, making it a generator-conditional
    safeguard rather than a high-volume filter.
  & Fig.~\ref{fig:abl2}, Table~\ref{tab:abl2_model} \\
3 & Retrieval is model-dependent: $+12.5$~pp gate compliance for one
    generator, no measurable effect for a second, and output collapse in a
    third.
  & Figs.~\ref{fig:abl3}, \ref{fig:camel} \\
\end{tabular}

\vspace{3pt}
\noindent Across gated configurations SNOMED density is flat
(29.71--29.86 terms/100w), and corpus-quality gains do not produce
commensurate improvement on real lung-cancer notes
(\S\ref{sec:res_tstr}).
\end{tcolorbox}
\subsection{Corpus Quality Across Ablation Conditions}
\label{sec:corpusquality}

Table~\ref{tab:corpus_quality} reports corpus-level quality metrics for all
five adapter conditions: generation yield, per-constraint pass rates (schema,
ontology, AJCC logic), length-normalized SNOMED\,CT density, and the number
of distinct SNOMED concepts in each admitted corpus. Three cross-cutting
observations follow.
\begin{table*}[!t]
\centering
\caption{Corpus quality summary by ablation condition. Yield = admitted /
generated. Schema, Onto., and Logic pass rates are computed over all generated
records. All entropy floors ($H_{T},H_{N} \geq 1.109$; $H_{M} \geq 0.554$)
pass in every condition. SNOMED density is length-normalized
(terms per 100 words) over admitted records. Unique = distinct SNOMED
concept IDs across all admitted records.}
\label{tab:corpus_quality}
\setlength{\tabcolsep}{5pt}
\begin{tabular}{@{}llccccccc@{}}
\toprule
\textbf{Adapter} & \textbf{Condition} &
\textbf{Gen.} & \textbf{Yield} &
\textbf{Schema} & \textbf{Onto.} & \textbf{Logic} &
\textbf{SNOMED/100w} & \textbf{Unique} \\
\midrule
A & Ungated            & 512 & 100.0\% & 70.1\% & 100.0\% & 79.9\% & 28.39 & 213 \\
B & Schema only        & 512 &  70.5\% & 70.5\% & 100.0\% & 78.1\% & 29.71 & 192 \\
C & Schema + Onto.     & 512 &  70.3\% & 70.3\% & 100.0\% & 77.0\% & 29.86 & 201 \\
D & Full $G(x)$, No-RAG  & 512 &  66.4\% & 71.1\% & 95.3\% & 77.1\% & 29.83 & 198 \\
E & Full $G(x)$ + RAG    & 448 &  69.2\% & 70.3\% & 83.3\% & 71.0\% & 29.52 & 191 \\
\bottomrule
\end{tabular}
\end{table*}
\begin{itemize}
    \item \textbf{Label diversity is preserved across all conditions:} Every corpus passes all three entropy floors, with measured diversity at the theoretical maxima (Appendix~\ref{app:diversity}). No entropy intervention is triggered at any phase across any adapter condition.
    \item \textbf{SNOMED density is insensitive to gate strictness:} Corpora B, C, and D show near-identical density (29.71, 29.86, and 29.83 terms per 100 words respectively). Enforcing progressively stricter gate constraints does not produce records with more ontological content; the gate filters for clinical validity, not vocabulary richness.
    \item \textbf{Model compliance is strongly bimodal:} GPT-4o and Llama-3.3-70B achieve near-perfect schema and logic compliance under gated conditions. ClinicalCamel-70B maintains 20--23\% schema compliance regardless of gate configuration or retrieval context---a property of the model, not the gate. These effects are reported per-model in each ablation below.
\end{itemize}

\subsection{RQ1 (Gate Necessity) --- Gated vs.\ Ungated Generation}
\label{sec:res_abl1}

\begin{figure*}[!t]
  \centering
  \includegraphics[width=0.8\textwidth]{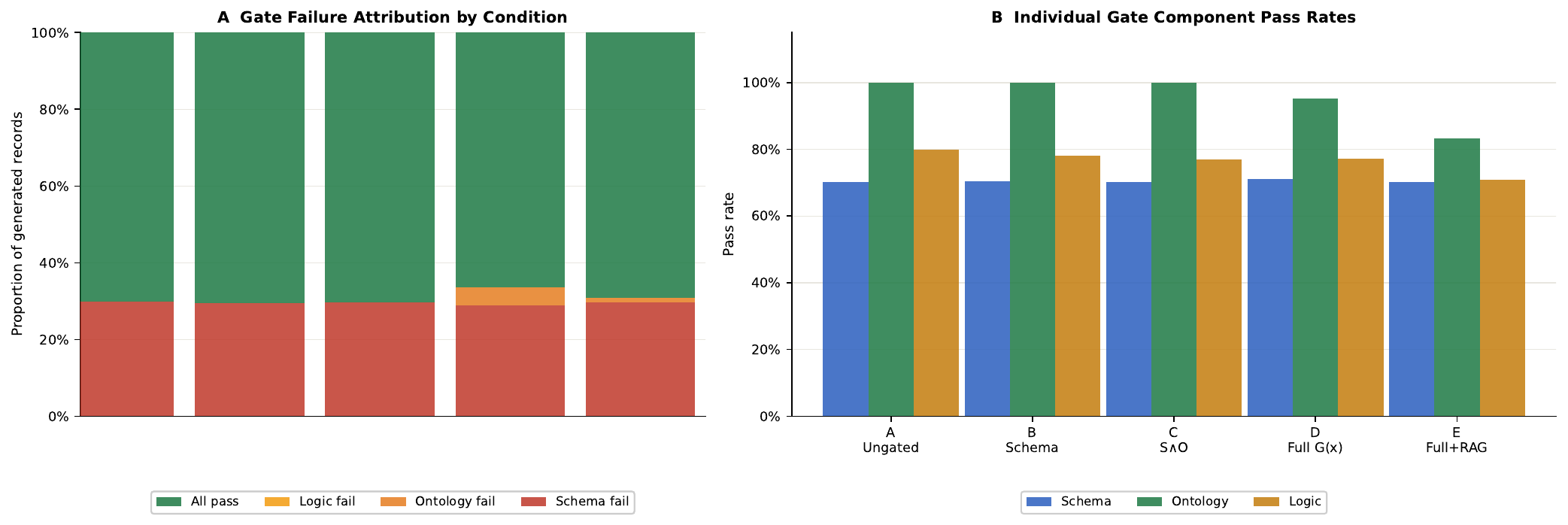}
  \caption{\textbf{Gate vs.\ No-Gate.}
\textit{(a)} Stacked gate failure attribution across all five conditions.
Column~A (Ungated) contains 29.9\% schema failures and 20.1\% logic failures
that would silently enter fine-tuning without the gate. Column~D (Full
$G(x)$) shows the residual failure structure after all three constraints are
applied; admitted records pass all three simultaneously. \textit{(b)}
Individual gate component pass rates. Ontology coverage reaches 100\% for
conditions A, B, and C by design (the schema mandates histology codes); the
full gate (D, E) applies all three constraints independently, and the
residual ontology failures at D are the 24 Llama records identified in
Fig.~\ref{fig:abl2}.}
  \label{fig:abl1}
\end{figure*}

Figure~\ref{fig:abl1} presents gate failure attribution and per-component
pass rates across all five conditions. The ungated corpus (Adapter~A)
admitted all 512 generated records but silently included substantial
clinical noise: 29.9\% of records failed schema validation and 20.1\%
violated AJCC clinical logic (Fig.~\ref{fig:abl1}a, column~A). Neither
failure is detectable through standard formatting inspection---they manifest
as missing JSON fields and logically impossible staging combinations
respectively. The full-gate corpus (Adapter~D) admitted 340
of 512 generated records (66.4\% yield), with every admitted record
satisfying all three constraints simultaneously. The gate therefore
eliminates one in three generated records and removes both detected categories
of clinical noise from the training corpus.

\textbf{The gate cost is asymmetric across generator models}
(Table~\ref{tab:abl1_model}). GPT-4o achieves 100\% gate compliance under
the full gate (128/128 records admitted) with no yield loss, reflecting
API-level JSON enforcement (\texttt{response\_format=\{type: json\_object\}})
that provides structural guarantees unavailable to locally-served models.
Llama-3.3-70B achieves 100\% schema and logic compliance in both conditions;
under the full gate its yield drops to 87.5\% (168/192 records), with all
24 rejected records failing the ontology constraint specifically.
ClinicalCamel-70B exhibits near-identical schema compliance ungated (20.3\%)
and under the full gate (22.9\%), confirming that its compliance limitation
is a model-level property orthogonal to gate configuration.

\begin{table}[!t]
\centering
\caption{Ablation~1 gate metrics by model. Schema, Onto., and Logic
are pass rates over all generated records.
Yield for the ungated condition is 100\% by construction (no gate
applied); for the full-gate condition it equals the fraction of records
satisfying all three constraints simultaneously.}
\label{tab:abl1_model}
\setlength{\tabcolsep}{4pt}
\begin{tabular}{@{}llccccc@{}}
\toprule
\textbf{Model} & \textbf{Condition} &
\textbf{Schema} & \textbf{Onto.} & \textbf{Logic} & \textbf{Yield} \\
\midrule
GPT-4o        & Ungated   & 100.0\% & 100.0\% & 100.0\% & 100.0\% \\
GPT-4o        & Full gate & 100.0\% & 100.0\% & 100.0\% & 100.0\% \\
\addlinespace[2pt]
Llama-3.3-70B & Ungated   & 100.0\% & 100.0\% & 100.0\% & 100.0\% \\
Llama-3.3-70B & Full gate & 100.0\% & 87.5\%  & 100.0\% & 87.5\%  \\
\addlinespace[2pt]
ClinicalCamel & Ungated   & 20.3\%  & 100.0\% & 46.4\%  & 100.0\% \\
ClinicalCamel & Full gate & 22.9\%  & 100.0\% & 39.1\%  & 22.9\%  \\
\bottomrule
\end{tabular}
\end{table}

Over admitted records, mean SNOMED density rises only slightly, from
28.39 (ungated) to 29.83 terms per 100 words under the full gate
($+5.1\%$). This modest increase
reflects the exclusion of low-density ClinicalCamel records---which dominate
the ungated corpus failures---rather than any vocabulary enrichment effect of
the gate itself.

\subsection{RQ2 (Constraint Attribution) --- Marginal Contribution of Gate Components}
\label{sec:res_abl2}

\begin{figure*}[!t]
  \centering
  \includegraphics[width=0.8\textwidth]{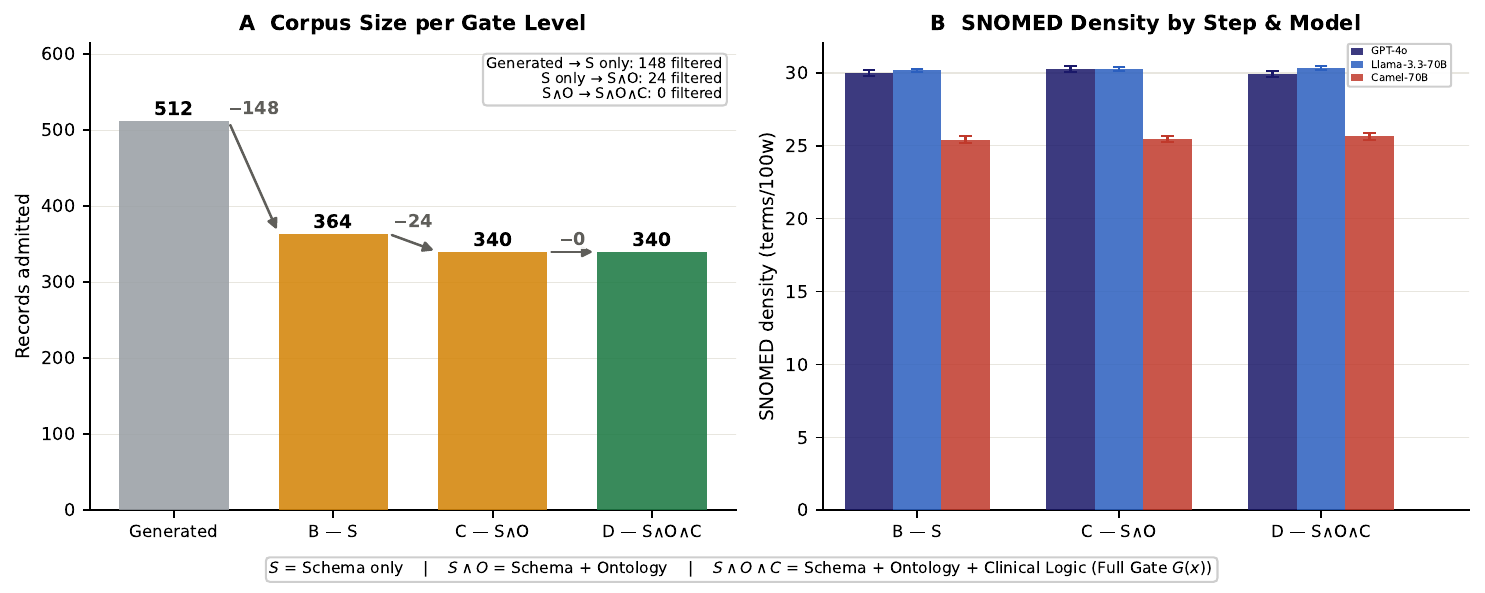}
  \caption{\textbf{Gate decomposition (within-corpus).}
\textit{(a)} Gate funnel within the full-gate corpus: of 512 generated
  records, schema admits 364; ontology grounding then removes 24, leaving
  340; and AJCC clinical-logic validation removes none. The marginal
  rejections---schema 148 (all ClinicalCamel), ontology 24 (all Llama),
  logic 0---identify schema as the load-bearing filter. \textit{(b)} SNOMED
  density by model across gate levels; flat for GPT-4o and Llama
  ($\approx$30 terms/100w), lower for ClinicalCamel ($\approx$25),
  confirming the gate filters validity, not vocabulary.
}
  \label{fig:abl2}
\end{figure*}

Figure~\ref{fig:abl2} shows the within-corpus gate funnel, the marginal
rejection attributable to each constraint, and SNOMED density across gate
levels. Computed within the full-gate corpus (512 generated records), schema
validation is the load-bearing filter: it admits 364 records and rejects
148, overwhelmingly from ClinicalCamel-70B. Of the 364 schema-passing
records, the ontology constraint removes a further 24---all from
Llama-3.3-70B---and the AJCC clinical-logic constraint removes none (Fig.~\ref{fig:abl2}a).

The zero marginal contribution of the logic constraint is robust. Applying a
substantially fuller AJCC consistency check (M1$\Rightarrow$Stage~IV,
M0$\not\Rightarrow$Stage~IV, Stage~I with N$\geq$1 or T$\geq$3, N3 below
Stage~IIIB) to every schema-passing record with machine-parseable staging
surfaces no clinically impossible assignments. The compliant generators
(GPT-4o, Llama-3.3-70B) do not produce AJCC contradictions, and the generator
that does (ClinicalCamel-70B) is removed by the schema constraint before
logic is evaluated. The clinical-logic gate is therefore a
\emph{generator-conditional} safeguard: the class of error it is designed to
catch---for example, M1 designation without a Stage~IV assignment, or
Stage~I co-occurring with N2/N3 involvement---does occur in principle, but no
such record reaches the logic check in this study because schema filtering
removes the only non-compliant generator first.

SNOMED density is flat across the three gate levels on the admitted-record
basis: 29.71, 29.86, and 29.83 terms per 100 words for B, C, and D
respectively (Fig.~\ref{fig:abl2}b). Enforcing the ontology constraint does
not enrich the vocabulary of admitted records; it standardizes a small
fraction whose generated concepts fall outside the controlled vocabulary. The
gate operates on clinical validity, not vocabulary richness.

By model, GPT-4o and Llama-3.3-70B maintain 100\% schema and logic compliance
at every gate level (Table~\ref{tab:abl2_model}). ClinicalCamel-70B's logic
pass rate over all generated records is low and roughly constant across gate
levels (41.7\%, 38.5\%, 39.1\%), but its logic-violating records also fail
schema, so they are removed by the schema constraint rather than contributing
marginal logic-gate rejections.

\begin{table}[!t]
\centering
\caption{Ablation~2 gate metrics by model across all three gate
levels. GPT-4o and Llama-3.3-70B maintain near-perfect compliance at
every level; ClinicalCamel-70B's compliance (21--23\%) is unchanged by which
constraints are active; its logic violations co-occur with schema failures,
so they are removed by the schema constraint rather than contributing
marginal logic-gate rejections.}
\label{tab:abl2_model}
\setlength{\tabcolsep}{4pt}
\begin{tabular}{@{}llcccc@{}}
\toprule
\textbf{Model} & \textbf{Gate level} &
\textbf{Schema} & \textbf{Onto.} & \textbf{Logic} & \textbf{Yield} \\
\midrule
GPT-4o        & $S$                       & 100.0\% & 100.0\% & 100.0\% & 100.0\% \\
GPT-4o        & $S \wedge O$              & 100.0\% & 100.0\% & 100.0\% & 100.0\% \\
GPT-4o        & $S \wedge O \wedge C$     & 100.0\% & 100.0\% & 100.0\% & 100.0\% \\
\addlinespace[2pt]
Llama-3.3-70B & $S$                       & 100.0\% & 100.0\% & 100.0\% & 100.0\% \\
Llama-3.3-70B & $S \wedge O$              & 100.0\% & 100.0\% & 100.0\% & 100.0\% \\
Llama-3.3-70B & $S \wedge O \wedge C$     & 100.0\% & 87.5\%  & 100.0\% & 87.5\%  \\
\addlinespace[2pt]
ClinicalCamel & $S$                       & 21.4\%  & 100.0\% & 41.7\%  & 21.4\%  \\
ClinicalCamel & $S \wedge O$              & 20.8\%  & 100.0\% & 38.5\%  & 20.8\%  \\
ClinicalCamel & $S \wedge O \wedge C$     & 22.9\%  & 100.0\% & 39.1\%  & 22.9\%  \\
\bottomrule
\end{tabular}
\end{table}

\subsection{RQ3 (Retrieval Conditionality) --- Retrieval-Augmented vs.\ Non-Augmented Generation}
\label{sec:res_abl3}

\begin{figure*}[!t]
  \centering
  \includegraphics[width=0.8\textwidth]{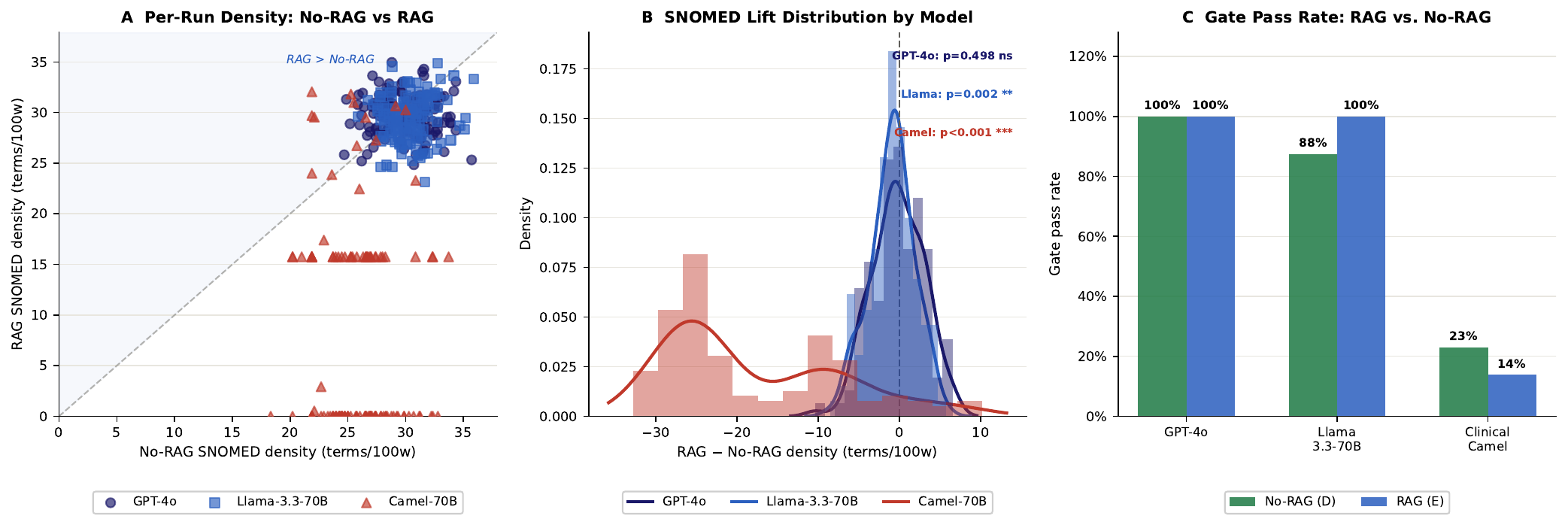}
  \caption{\textbf{RAG vs.\ No-RAG.}
  \textit{(a)} Per-run SNOMED density scatter. GPT-4o and Llama points
  cluster symmetrically around the diagonal; ClinicalCamel triangles
  collapse to the $y{=}0$ and $y{=}15.75$ floor under RAG.
  \textit{(b)} Distribution of per-run density delta (RAG $-$ No-RAG) by
  model. GPT-4o centered near zero (two-sided $p = 0.498$, ns); Llama a small
  but significant negative shift ($p = 0.002$); ClinicalCamel strongly
  negative ($p < 0.001$), reflecting retrieval collapse.
  \textit{(c)} Gate pass rate by model. RAG lifts Llama gate compliance
  from 87.5\% to 100\%; ClinicalCamel falls from 22.9\% to 13.8\%.}
  \label{fig:abl3}
\end{figure*}

Figure~\ref{fig:abl3} presents the RAG ablation results. The retrieval intervention exhibits strongly model-dependent behavior, motivating examination of generator-specific effects before evaluating downstream utility.

\textbf{Overall}, the RAG corpus (Adapter~E) shows a gate pass rate of
69.2\% versus 66.4\% for No-RAG (Adapter~D, $+2.8$~pp). On the
admitted-record basis, mean SNOMED density (29.52 vs 29.83) and unique
concept coverage (191 vs 198) are essentially unchanged: retrieval does not
enrich the corpus a model trains on. The aggregate effect of retrieval is
instead dominated by a single generator, examined below. (Measured over
\emph{all} generation attempts---including rejected and collapsed runs---mean
density falls from 27.05 to 23.18 terms per 100 words; this all-generation
drop is driven entirely by the ClinicalCamel collapse, not by any change to
admitted records.)

\textbf{GPT-4o} shows a mean per-run density delta of $-0.20$ terms per 100
words (Mann-Whitney, two-sided, $p = 0.498$, ns), with a gate pass rate of
100\% in both conditions. RAG adds no measurable signal for a model already
operating at ceiling on every metric.

\textbf{Llama-3.3-70B} exhibits the most consequential RAG effect: gate
pass rate increases from 87.5\% (No-RAG) to 100.0\% (RAG), a
$+12.5$~pp lift (Fig.~\ref{fig:abl3}c). The 24 records that Llama fails
in the No-RAG condition all fail the ontology constraint; retrieved PubMed
context provides sufficient SNOMED\,CT grounding to bring these records into
compliance. The mean per-run density delta is $-0.92$ terms per 100 words
(Mann-Whitney, two-sided, $p = 0.002$): retrieval slightly \emph{lowers}
per-run density even as it lifts gate compliance, confirming that RAG helps
Llama pass the gate by admitting previously-failing records rather than by
enriching the vocabulary of already-admitted ones.

\begin{figure*}[!t]
  \centering
  \includegraphics[width=0.8\textwidth]{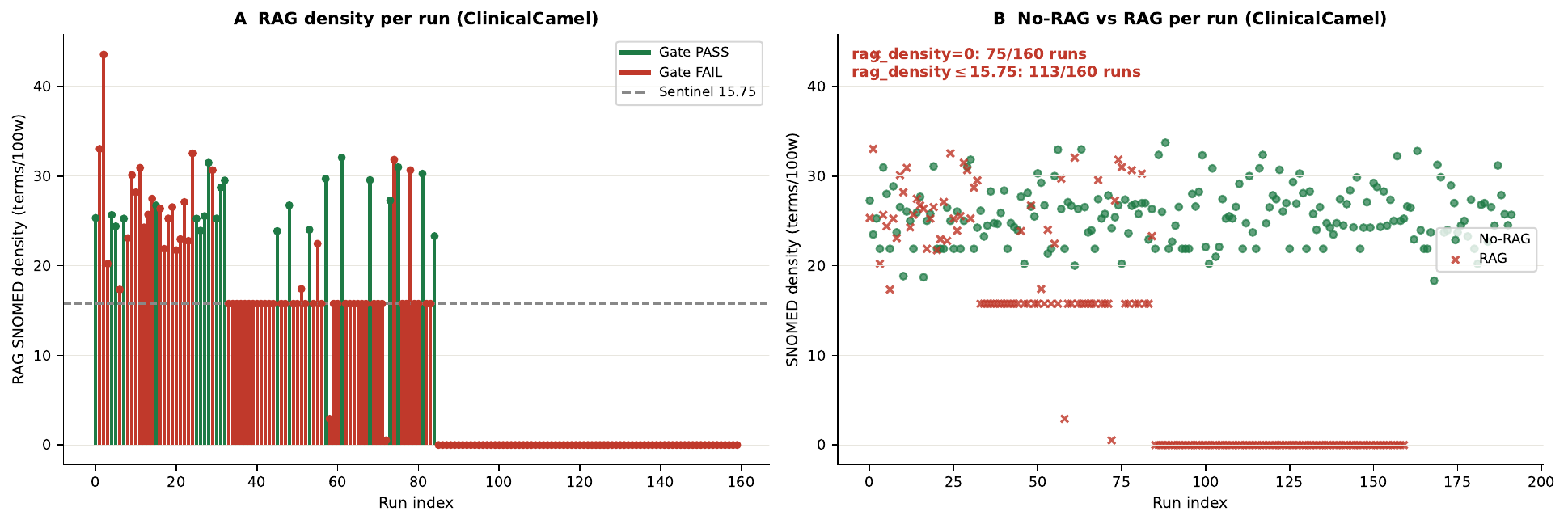}
  \caption{\textbf{ClinicalCamel-70B RAG Collapse.}
  \textit{(a)} RAG SNOMED density per run colored by gate outcome
  (green = PASS, red = FAIL). Runs 0--84 show mixed outcomes with
  densities around and below 15.75; from run~85 onward, density collapses to
  exactly 0.0 in nearly all runs, corresponding to complete output failure.
  \textit{(b)} No-RAG (circles) versus RAG (crosses) density paired by run.
  No-RAG density is stable throughout (20--33 terms/100w); RAG density
  disintegrates in the second half of the run sequence. 75/160 runs produce
  zero SNOMED output and 113/160 produce density $\leq 15.75$ under RAG.}
  \label{fig:camel}
\end{figure*}

\textbf{ClinicalCamel-70B} is the source of the aggregate density drop.
In the No-RAG condition ClinicalCamel produces a mean SNOMED density of
25.64 terms per 100 words and a gate pass rate of 22.9\%, consistent with
its behavior across all other conditions. Under RAG, 75 of 160 runs (46.9\%) produce a SNOMED density of exactly 0.0,
and 113 of 160 runs (70.6\%) produce a density at or below 15.75
(Mann-Whitney $p < 0.001$). Gate pass rate falls from 22.9\% to 13.8\%; 138
of 160 RAG runs fail the gate entirely. This is a retrieval collapse: injecting retrieved context into
ClinicalCamel's already-fragile generation format causes systematic output
disintegration beginning around run~85, where density transitions from
mixed outcomes to near-universal zero output. Figure~\ref{fig:camel}
documents this collapse at the run level.

\subsection{Downstream Utility: Train-on-Synthetic, Test-on-Real Evaluation}
\label{sec:res_tstr}

\begin{figure*}[!t]
  \centering
  \includegraphics[width=0.8\textwidth]{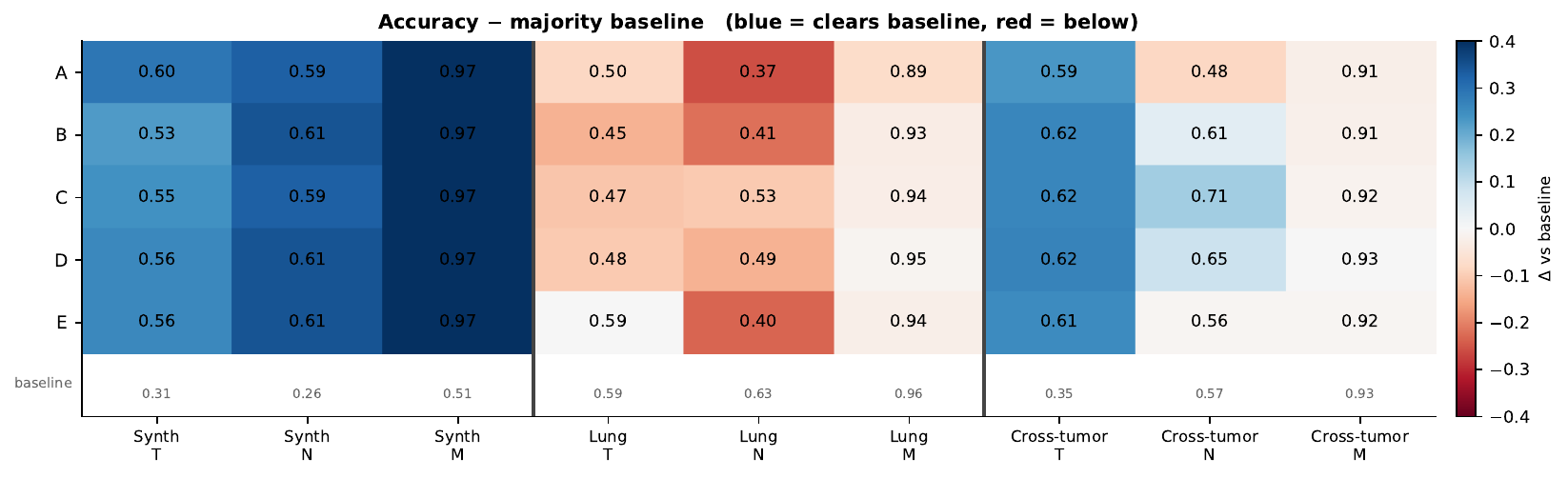}
\caption{\textbf{Downstream TSTR by axis and cohort.} Cell color is accuracy
minus the majority-class baseline (blue clears the baseline, red falls below);
the printed value is raw accuracy, and the bottom row gives the baseline per
column. In-distribution (Synthetic) every adapter clears the baseline on all
three axes; on real TCGA accuracy collapses toward the majority class,
especially lung N-stage. On Cross-Tumor, ontology-aware adapters C and D most
decisively clear the N-stage baseline. Per-axis macro-F1 is
reported in Table~\ref{tab:tstr_f1}.}
  \label{fig:tstr}
\end{figure*}

Figure~\ref{fig:tstr} summarizes downstream T/N/M staging performance across
all five adapter conditions. Each adapter is fine-tuned on its own admitted
synthetic corpus---1,883 admitted records in total across conditions
A--E---and evaluated under the train-on-synthetic, test-on-real (TSTR)
protocol on three benchmarks of increasing distributional distance: a
held-out synthetic test set ($n=62$), TCGA Lung ($n=737$), and TCGA
Cross-Tumor ($n=3{,}161$). Each adapter produces a T-, N-, and M-stage
prediction for every note---3,960 evaluation notes per adapter, 19,800
note-level predictions across the five conditions---with per-axis scoring
restricted to notes carrying a valid gold label for that axis.

The following patterns were observed:

\textbf{First, all adapter conditions perform well in-distribution.}
Across the held-out synthetic benchmark, accuracy and macro-F1 remain
consistently high regardless of gate configuration. Differences between
adapters are modest relative to the real-world benchmarks, indicating
that every synthetic corpus contains sufficient signal to support
learning within the distribution from which it was generated.

\textbf{Second, strong synthetic performance does not translate to TCGA
Lung.} Despite substantial differences in corpus quality between the five generation conditions, no adapter exceeds the
majority-class baseline on any axis of the lung pathology benchmark. Improvements in
schema validity, ontology compliance, and clinical-logic consistency
produce cleaner synthetic corpora, but these gains do not yield
commensurate improvements in lung-cancer generalization. The dominant
effect is therefore the transition from synthetic generation outputs to
real pathology reports rather than differences among gate configurations.

\textbf{Third, the largest between-adapter differences appear on the
most heterogeneous benchmark.} On TCGA Cross-Tumor every adapter clears the T-stage baseline decisively
(0.59--0.62 against a baseline of 0.35), the one benchmark where the
synthetic corpora transfer. Adapters~C and~D additionally record the
strongest N-stage performance, a separation not visible on the synthetic
benchmark and only weakly apparent on TCGA Lung. With a single training
seed per condition this gap is not separated from seed variance; we report
it as an observation consistent with ontology grounding carrying portable
signal across disease contexts, not as an established effect.

A consistent pattern across all conditions is the discrepancy between
accuracy and macro-F1 for M-stage prediction. While M-stage accuracies
appear high, macro-F1 remains substantially lower, reflecting the strong
class imbalance of the evaluation sets and the predominance of M0 cases.
Consequently, M-stage accuracy should not be interpreted as evidence of
superior metastatic classification performance.

\section{Discussion}
\label{sec:discussion}
\subsection{A Ranked Answer to the Component Attribution Question}
\label{sec:disc_ranking}

The three ablations provide direct answers to RQ1, RQ2, and RQ3 while
also allowing the pipeline components to be ranked by their marginal
contribution to corpus quality.

Schema validation is the load-bearing constraint. Within a single corpus it
rejects 148 of 512 generated records---almost all from the one generator
(ClinicalCamel-70B) whose structured-output compliance is low---and the
records it removes are also the ones that carry AJCC logic violations.
Ontology grounding contributes a small, generator-specific filtering
increment (24 Llama records) and otherwise standardizes vocabulary. AJCC
clinical-logic validation contributes no marginal filtering under the
conditions of this study, because the only logic-violating generator is
already excluded by schema; it functions as a generator-conditional safeguard
that would become load-bearing for any generator emitting schema-valid but
clinically impossible records.

This ranking has a direct practical implication. A practitioner building a
similar pipeline should implement schema validation first as a minimum
viable quality floor, retain clinical-logic validation as a generator-conditional safeguard, treat ontology grounding as a normalization step
rather than a filter, and evaluate retrieval augmentation per-generator
before enabling it. The ordering is not arbitrary---it reflects the marginal
contribution of each component to corpus clinical validity, measured
independently under controlled conditions.

\subsection{The Gate Filters Validity, Not Vocabulary}
\label{sec:disc_validity}

A consistent finding across all three ablations is that SNOMED\,CT density
is insensitive to gate strictness. Corpora B, C, and D show densities of
29.71, 29.86, and 29.83 terms per 100 words---essentially flat, with any
residual variation attributable to model mix rather than gate configuration.
This directly falsifies the hypothesis that the gate enriches vocabulary by
excluding low-density records.

The practical implication is that SNOMED density is not a sufficient proxy
for corpus quality. A record can carry 35 SNOMED terms per 100 words while
assigning M1 staging without a Stage~IV label---ontologically rich and
clinically invalid simultaneously. Ontological richness and clinical
correctness are independently evaluable properties, and conflating them
leads to a false sense of corpus quality. Pipelines that report only
vocabulary coverage metrics as quality indicators are measuring the wrong
thing: a high SNOMED density confirms that admitted records use recognized
clinical terminology, but says nothing about whether the clinical reasoning
those terms encode is internally consistent.

It also separates the two constraints: $O$ does standardization, not
filtering, while $C$ is the mechanism that would catch logical errors. The
ontology gate is not redundant, but its contribution is qualitatively
different from the logic gate's.

\subsection{RAG as Gate Compliance Mechanism, Not Vocabulary Enrichment}
\label{sec:disc_rag}

The RAG ablation resolves a question left open by the vocabulary enrichment
finding from prior work in this space: does SNOMED density enrichment from
retrieval translate into downstream training signal, or does the gate already
enforce sufficient ontological grounding to make retrieval redundant? The
answer is model-conditional, and the model-dependency is sharp enough to
constitute a design warning rather than a nuanced finding.

For Llama-3.3-70B, RAG functions as a gate compliance mechanism rather than
a vocabulary enrichment tool. The 24 records that fail the ontology
constraint in the No-RAG condition are resolved when retrieved PubMed
context provides sufficient SNOMED grounding, lifting gate pass rate from
87.5\% to 100\%. Crucially, density does not increase on already-admitted
records: RAG admits records that would have been excluded, not richer
versions of records that would have been admitted anyway. The effect is
yield improvement, not quality improvement of individual records.

For GPT-4o the gate is already satisfied at 100\%, so retrieval has no
failures to resolve and contributes nothing measurable.

For ClinicalCamel-70B, RAG is actively harmful. Injecting retrieved context
into a generation format that is already fragile causes systematic output
disintegration. The transition is abrupt: the first 85 runs show mixed
outcomes with densities around and below 15.75; from run 85 onward, 75
consecutive runs produce zero SNOMED output. This is not a gradual
degradation---it is a collapse, and its abruptness suggests a context-length
or prompt-template threshold rather than a gradual interference effect.

The practical lesson is that retrieval augmentation should be treated as a
per-generator decision rather than a pipeline default. The gate provides a
natural diagnostic: if a generator's No-RAG gate compliance is already at
ceiling, RAG adds no value and may introduce failure modes. If compliance
is below ceiling, RAG may resolve specific constraint failures---but only
if the generator can integrate retrieved context without format
disintegration, which must be verified empirically before deployment.

\subsection{Domain Pre-training and Format Compliance Are Orthogonal}
\label{sec:disc_camel}

ClinicalCamel-70B was included to test whether domain-adaptive clinical
pre-training confers an advantage for structured oncology generation. It
does not: schema compliance holds at 20--23\% across every gate
configuration and retrieval condition---roughly one valid record in five
runs regardless of what constraints are active.

This finding challenges an assumption that is common in clinical NLP: that
a model pre-trained on clinical text will produce better-structured clinical
output than a general-purpose model. The evidence here suggests that
domain knowledge and structured format compliance are largely orthogonal
capabilities. ClinicalCamel-70B demonstrably encodes clinical knowledge---
its SNOMED density on the records it does produce is comparable to
Llama-3.3-70B---but its instruction-following capability and JSON format
compliance are insufficient for structured generation tasks regardless of
that knowledge.

The broader implication is that generator selection should be evaluated
primarily on format compliance and instruction-following fidelity, not on
clinical pre-training provenance. The gate enforces the same quality floor
regardless of generator, but generator choice determines yield: domain
pre-training is neither necessary nor sufficient for the compliance that
matters here.

\subsection{Toward Downstream Transferability}
\label{sec:disc_transfer}

The downstream results locate the next problem precisely. The gate secures the validity of admitted records, but on real pathology
reports it is the synthetic-to-real
gap, not the gate configuration, that dominates T-stage performance. The
generated corpus, though clinically valid, is narrow: the fixed 32-cell TNM
seeding grid and templated prompts were chosen for controlled, reproducible
ablation, but they bound the lexical and structural variety of the reports a
model trains on, and a clean but narrow corpus transfers imperfectly to the
full distribution of real pathology narratives. A second limitation concerns
yield rather than validity---the records the gate discards are errors the
generator still produces, rejected at a direct cost in corpus size without
any change to generator behavior. A third observation is where transfer did
succeed: ontology-aware filtering cleared the majority-class baseline on
cross-tumor nodal staging, suggesting that ontological grounding carries
portable signal across disease contexts.
The central limitation is that each condition was trained once. The
downstream comparisons therefore establish that a single gated run did not
outperform a single ungated run on real lung notes; they do not establish
that gating cannot improve transfer. Distinguishing the two requires
replication across seeds, which at roughly one minute of fine-tuning per
adapter is inexpensive and is the first thing we would add.
\subsection{Future Directions}
\label{sec:future}
Three directions follow;
The first is to generate
richer notes rather than to filter harder: prompting from real pathology
exemplars and institution-specific reporting styles, covering TNM
combinations beyond the 32-cell grid used here, and generating a patient's
record over time rather than a single snapshot, so that a model sees how
stage is revised across a course of treatment. The second is to move the
ontology and AJCC checks into decoding, so that the generator produces fewer
invalid records instead of discarding them afterward. This raises yield, not
validity: the corpus would be no more valid than the gate already makes it.
The third is to widen the subject domain to tumor types whose AJCC rules are
built differently---bladder, colorectal, head and neck---and to widen $C$
from the commonly violated rules implemented here to the full staging logic.
Generator-specific robustness assessment was beyond the scope of this study.

\section{Conclusion}
\label{sec:conclusion}

This paper measured which components of a neuro-symbolic synthetic clinical
data pipeline are load-bearing, and by how much, through five matched
adapter conditions and three controlled ablations.

Four findings follow. Schema validation does most of the filtering, and
because the only logic-violating generator fails schema first, it also
removes the impossible-staging records before any later check reaches them.
Clinical-logic validation removes nothing marginal here; it is a
generator-conditional safeguard, not a high-volume filter. Ontology
grounding standardizes rather than filters: SNOMED density is flat across
gate levels. Retrieval augmentation is model-conditional, resolving ontology
failures for one generator, adding nothing for a second, and collapsing the
output of a third.

These results support a general design claim. Formalized clinical rule
systems---staging manuals, coding taxonomies, drug-interaction databases---can
be encoded as generation-time admission constraints rather than applied as
post-hoc evaluation labels, which makes hallucination detection binary,
auditable, and reproducible at the cost of reduced yield.

They also locate the remaining leverage. Across real TCGA pathology reports,
gated and ungated adapters were largely indistinguishable on T-stage. The
gate is where validity is secured. Downstream utility is won on the
generation side: producing more clinical language, more varied in vocabulary
and structure, and longitudinal rather than snapshot---records that follow a
patient across a course of treatment, where stage is revised as evidence
accumulates. Such a corpus still needs the gate. It also needs to be far
wider than the one measured here.

\section*{Data and Code Availability}
Code, generation logs with per-record gate flags, and adapter checkpoints
are available at
\url{https://github.com/LGChalla/Synthetic-Data-Ablation-Oncology}.
Evaluation cohorts derive from The Cancer Genome Atlas, accessed under the
NIH Genomic Data Sharing Policy; all generated records are synthetic and no
protected health information was used.

\section*{Ethical Considerations}
The \textit{rigid.v3} schema mandates a
\texttt{health\_equity\_factors} field, so demographic and equity
attributes are model-generated rather than sampled from a reference
population, and the gate does not constrain them: it validates staging
logic and ontology coverage, not whether generated associations between
demographic attributes and stage, histology, or treatment intent reproduce
biases in the generators' pre-training data. We did not audit for such
associations, and corpora produced this way should not be used to study
disparities without one. This is the construct problem the paper documents
for SNOMED density---a constraint certifies only what it is defined
over---and extending admission constraints to demographic plausibility
follows directly from the mechanism described here.

\appendices

\section{Entropy Floor Derivation and Label Diversity}
\label{app:diversity}

Entropy floors are set at 80\% of maximum entropy per TNM dimension. For
T and N ($k=4$): $H_{\max} = \ln 4 = 1.386$ nats, floor $= 1.109$. For M
($k=2$): $H_{\max} = \ln 2 = 0.693$ nats, floor $= 0.554$. The floors are
checked twice---after generation, on the admitted corpus before the
train/test split, and before training, on the training partition after
stratified splitting---with fine-tuning aborted if any dimension is
single-class. No entropy intervention was triggered at either checkpoint
in any of the five adapter conditions.

\begin{table}[!ht]
\centering
\caption{Label diversity (Shannon entropy, nats) of the admitted corpus per
TNM axis. All conditions sit at the theoretical maxima ($\ln4=1.386$ for
T,N; $\ln2=0.693$ for M), well above the 80\% floors, because the 32-cell
seed grid balances labels by construction.}
\label{tab:entropy}
\setlength{\tabcolsep}{6pt}
\begin{tabular}{@{}lccc@{}}
\toprule
\textbf{Adapter} & \textbf{$H_T$} & \textbf{$H_N$} & \textbf{$H_M$} \\
\midrule
A--E (all) & 1.39 & 1.39 & 0.69 \\
\midrule
Floor (0.8$\times$max) & 1.109 & 1.109 & 0.554 \\
\bottomrule
\end{tabular}
\end{table}
\section{Per-Axis Macro-F1 Scores}
\label{app:macrof1}
Table~\ref{tab:tstr_f1} reports per-axis macro-F1 for all five adapters
under the TSTR protocol, complementing the accuracy heatmap in
Fig.~\ref{fig:tstr}.
\begin{table}[!ht]
\centering
\caption{Per-axis macro-F1 for all five adapters under TSTR, across the
three evaluation cohorts. Macro-F1 weights each class equally, making it
sensitive to minority-class recovery in a way aggregate accuracy is
not---particularly for M-stage, where the M0 majority inflates accuracy
while macro-F1 stays low.}
\label{tab:tstr_f1}
\small
\setlength{\tabcolsep}{4pt}
\begin{tabular}{@{}lccccccccc@{}}
\toprule
& \multicolumn{3}{c}{\textbf{Synthetic}}
& \multicolumn{3}{c}{\textbf{TCGA Lung}}
& \multicolumn{3}{c}{\textbf{Cross-Tumor}} \\
\cmidrule(lr){2-4}\cmidrule(lr){5-7}\cmidrule(lr){8-10}
\textbf{Adapter} & T & N & M & T & N & M & T & N & M \\
\midrule
A & 0.56 & 0.60 & 0.97 & 0.41 & 0.33 & 0.53 & 0.56 & 0.53 & 0.67 \\
B & 0.50 & 0.61 & 0.97 & 0.40 & 0.33 & 0.59 & 0.58 & 0.56 & 0.66 \\
C & 0.48 & 0.59 & 0.97 & 0.41 & 0.41 & 0.59 & 0.59 & 0.64 & 0.67 \\
D & 0.50 & 0.62 & 0.97 & 0.42 & 0.39 & 0.58 & 0.59 & 0.60 & 0.66 \\
E & 0.53 & 0.61 & 0.97 & 0.44 & 0.33 & 0.58 & 0.57 & 0.55 & 0.69 \\
\bottomrule
\end{tabular}
\end{table}
\section{QLoRA Configuration}
\label{app:qlora}

All five adapters (A--E) are fine-tuned from the same base model using
identical QLoRA hyperparameters~\cite{hu2022lora,dettmers2023qlora}
(Table~\ref{tab:qlora}), ensuring
performance differences across adapter conditions are attributable to corpus
quality rather than training procedure. Training seed 42 is applied identically across all adapters, so differences
between adapters are not attributable to seed selection. A single seed per
condition does not, however, bound seed-to-seed variance itself.

\begin{table}[!htbp]
\centering
\caption{QLoRA fine-tuning configuration, identical across all five
adapters (A--E).}
\label{tab:qlora}
\small
\setlength{\tabcolsep}{5pt}
\renewcommand{\arraystretch}{1.2}
\begin{tabular}{@{}>{\raggedright\arraybackslash}p{2.5cm}
                   >{\raggedright\arraybackslash}p{\dimexpr\columnwidth-2.5cm-2\tabcolsep\relax}@{}}
\toprule
\textbf{Setting} & \textbf{Value} \\
\midrule
\multicolumn{2}{@{}l}{\textit{Base model and Quantization}} \\
Base model        & \texttt{meta-llama/}\allowbreak\texttt{Meta-Llama-3-8B-Instruct} \\
quantization      & 4-bit NF4, double quantization \\
Compute dtype     & bfloat16 \\
\midrule
\multicolumn{2}{@{}l}{\textit{LoRA}} \\
Rank ($r$)        & 16 \\
Scaling ($\alpha$) & 32 \\
Dropout           & 0.05 \\
Target modules    & \texttt{q\_proj}, \texttt{k\_proj}, \texttt{v\_proj}, \texttt{o\_proj} \\
Trainable params  & 13{,}631{,}488 / 8{,}043{,}892{,}736 (0.1695\%) \\
\midrule
\multicolumn{2}{@{}l}{\textit{Optimization}} \\
Epochs            & 3 \\
Learning rate     & $2 \times 10^{-4}$, cosine decay \\
Batch size        & 2 per device \\
Grad.\ accum.     & 4 (effective batch size 8) \\
Warmup steps      & 10 \\
Optimizer         & \texttt{paged\_adamw\_8bit} \\
Precision         & fp16 mixed \\
Max seq.\ length  & 1{,}024 tokens \\
Random seed       & 42 (all adapters) \\
\midrule
\multicolumn{2}{@{}l}{\textit{Hardware and runtime}} \\
GPU               & Single NVIDIA H200 \\
Training time     & ${\approx}60$\,s per adapter; ${<}5$\,min total \\
\bottomrule
\end{tabular}
\end{table}

\section{Prompt Details}
\label{app:prompts}

This appendix documents the two prompt families used in the pipeline. The
structured-generation prompt conditions each generator on a single TNM seed
drawn from the 32-cell grid (Section~\ref{sec:generation}) and instructs it
to return a complete record conforming to the \textit{rigid.v3} schema; in
the RAG condition (Adapter~E), the top-$k$ MedCPT-retrieved PubMed abstracts
are prepended to this template as grounding context preceding the
instruction. The extraction prompt elicits TNM staging from real clinical
notes under the TSTR protocol.

\begin{tcolorbox}[promptbox,
    title=\textbf{Structured Generation Prompt (\textit{rigid.v3})}]
\footnotesize\ttfamily\raggedright
You are a clinical oncology data generator. Generate one synthetic
non-small-cell lung cancer patient record as a single JSON object that
conforms exactly to the rigid.v3 schema. The record must be internally
consistent with AJCC 8th Edition staging. Seed: T=\{T\}, N=\{N\}, M=\{M\}.
Required fields: tnm\_staging, histology (ICD-O-3 and SNOMED CT codes),
molecular\_drivers, demographics, imaging\_findings, treatment\_modalities,
health\_equity\_factors, and narrative. Return only the JSON object with no
commentary.
\end{tcolorbox}

\begin{tcolorbox}[promptbox,
    title=\textbf{TSTR Extraction Prompt}]
\footnotesize\ttfamily\raggedright
You are a clinical data extractor. Read the clinical note and extract the
TNM staging. Return a strictly formatted JSON object with keys 'T', 'N', and
'M'. Always use the full prefixed format (e.g., T2, N0, M0). If a value is
not found, use 'Unknown'.
\end{tcolorbox}

\end{document}